\documentclass[letterpaper, 10 pt, conference]{ieeeconf}    

\IEEEoverridecommandlockouts                              

\usepackage{epsfig} 
\usepackage{times} 
\usepackage{amsmath} 
\usepackage{amssymb}  
\usepackage{subfigure}
\usepackage{graphics}
\usepackage{graphicx}
\usepackage{color}
\usepackage{multirow}
\usepackage{booktabs}
\usepackage{bigstrut}
\usepackage{hyperref}
\usepackage{color}
\usepackage{makecell}
\usepackage[table]{xcolor}
\usepackage{rotating}
\usepackage{bbding}
\usepackage[font=small]{caption}

\title{\LARGE \bf
DEVO: Depth-Event Camera Visual Odometry\\ in Challenging Conditions   
}

\author{Yi-Fan Zuo$^{1*}$, Jiaqi Yang$^{2*}$, Jiaben Chen$^{2}$, Xia Wang$^{1}$, Yifu Wang$^{2\dagger}$ and Laurent Kneip$^{2\dagger}$
\thanks{$^{*}$ indicates equal contribution, $^{\dagger}$ indicates corresponding author}
\thanks{$^{1}$ Key Laboratory of Optoelectronic Imaging Technology and Systems, Ministry of Education, School of Optics and Photonics, Beijing Institute of Technology, Beijing 100081, China.
$^{2}$Mobile Perception Lab, ShanghaiTech University; L. Kneip is also with the Shanghai Engineering Research Center of Intelligent Vision and Imaging.}%
}
        
\let\oldtwocolumn\twocolumn
\renewcommand\twocolumn[1][]{
     \oldtwocolumn[{#1}{
     \begin{center}
     \centering
     \includegraphics[width=\textwidth]{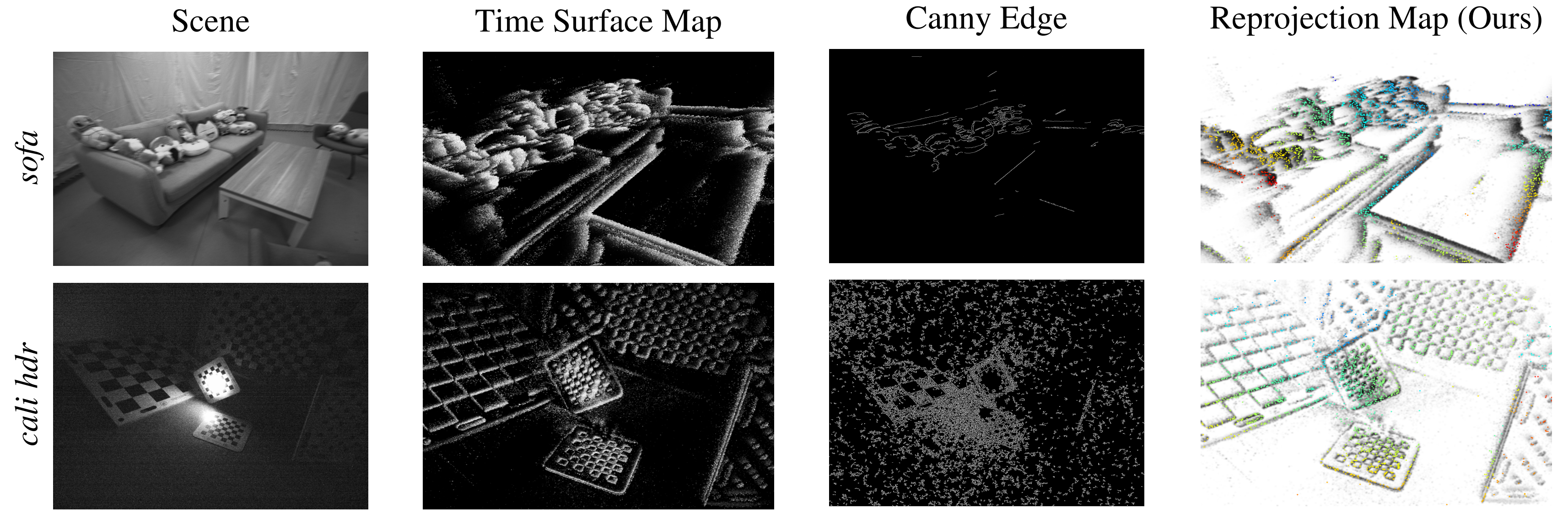}
     \captionof{figure}{\emph{Challenging scenarios and results}. 
	 \textit{sofa} is captured under high dynamics. \textit{cali hdr} is a challenging illumination scene with highly self-similar texture. Column 1: Example RGB images. Column 2: Corresponding time-surface map. Column 3: Canny edge detections. Column 4: Reprojected depth points.}
\end{center}
     }]
}

\begin{document}

\maketitle
\thispagestyle{empty}
\pagestyle{empty}

\begin{abstract}
We present a novel real-time visual odometry framework for a stereo setup of a depth and high-resolution event camera. Our framework balances accuracy and robustness against computational efficiency towards strong performance in challenging scenarios. We extend conventional edge-based semi-dense visual odometry towards time-surface maps obtained from event streams. Semi-dense depth maps are generated by warping the corresponding depth values of the extrinsically calibrated depth camera. The tracking module updates the camera pose through efficient, geometric semi-dense 3D-2D edge alignment. Our approach is validated on both public and self-collected datasets captured under various conditions. We show that the proposed method performs comparable to state-of-the-art RGB-D camera-based alternatives in regular conditions, and eventually outperforms in challenging conditions such as high dynamics or low illumination.
\end{abstract}

\section{Introduction}

Real-time localization and 3D mapping are increasingly important tasks to be solved in many emerging technologies such as robotics, intelligent transportation, and intelligence augmentation. Owing to their small scale and affordability, cameras are often considered as an exteroceptive sensor in such applications. Despite being attractive, pure vision-based solutions still lack robustness in more challenging conditions~\cite{fuentes2015visual,cadena2016past}, and are therefore often complemented by additional sensors such as an Inertial Measurement Unit (IMU), wheel encoders, or a depth camera. Especially the addition of the latter has been highly popular in indoor applications since the advent of consumer-grade RGB-D cameras in 2010 (e.g. Microsoft Kinect). RGB-D cameras provide high frequency and high resolution depth images, which significantly improves accuracy and robustness of monocular visual odometry and SLAM methods~\cite{steinbrucker2011real, endres2013rgbd, kerl2013robust, henry2014rgb, zhou2018canny}. However, most RGB-D camera solutions still rely on sparse feature extraction or photometric image alignment, which is why they cannot handle challenging conditions such as highly dynamic motion or low illumination. KinectFusion~\cite{newcombe2011kinectfusion}, relies exclusively on depth images, but the method is power hungry and demands high frame-rate depth images as well as GPU resources for the real-time execution of depth fusion and ICP algorithms.


The present work introduces a fresh approach to depth camera-supported indoor visual odometry (VO) for a power-efficient handling of challenging conditions such as high dynamics or low illumination. Our core idea consists of exchanging the RGB camera against a Dynamic Vision Sensor, which pairs high dynamic range (HDR) with high temporal resolution. The basic functionality of a DVS---also called an event camera---as well as its advantages and challenges are well explained in the original work of Brandli~et~al.~\cite{brandli2014240} or the recent survey by Gallego~et~al.~\cite{gallego2019event}. Our method relies on an approach similar to Canny-VO~\cite{zhou2018canny} and extracts edges from the event stream, assigns depth from the depth camera, and registers subsequent views by 3D-2D edge alignment. The ability of event cameras to see in almost complete darkness paired with their high temporal resolution lead to excellent performance in the above mentioned challenging cases. At the same time, our method maintains high computational and energy efficiency owing to potentially low depth camera frame rates as well as semi-dense processing. Our contributions are as follows:
%
%
\begin{itemize}
    \item We present DEVO, a novel visual odometry framework for a hybrid stereo setup of a depth and an event camera.
    \item The approach relies on thresholded time-surface maps for edge detection and semi-dense depth map extraction.
    \item Our method handles 6-DoF motion estimation, and we demonstrate high efficiency and successful operation in all conditions.
\end{itemize}
  
Our results are obtained on self-collected high-resolution RGB-D-event indoor datasets with ground truth captured by an external motion tracking system. Further tests are conducted on larger scale outdoor datasets where depth is obtained from a LiDAR scanner. A thorough comparison against state-of-the-art event-based and RGB-D based visual odometry frameworks proves that DEVO achieves high quality, continuous visual localization results and eventually outperforms alternative methods in challenging conditions.

\section{Related work}\label{sec:relatedWork}
Next, we review classical RGB-D camera approaches as well as both pure and multi-sensor, event-based visual odometry solutions.

\textbf{RGB-D camera-based solutions:}
The most straightforward solutions to RGB-D camera-based VO use only depth information \cite{newcombe2011kinectfusion,whelan2016elasticfusion}. While they may potentially operate in dark environments, they require dense depth image processing at high frame rate, and therefore require high energy and computation resources (e.g. GPU). Approaches that also rely on images~\cite{steinbrucker2011real, endres2013rgbd, kerl2013robust, henry2014rgb} often perform dense photometric alignment, and thus still depend on exhaustive parallel computing. They furthermore have the disadvantage of degrading in challenging visual conditions (e.g. blur, low illumination). Most related to our method are approaches relying on sparsified, semi-dense depth maps \cite{zhou2018canny, schenk2019reslam}. They have large convergence basins, stability under illumination changes, and high computational efficiency. Nonetheless, they still depend on intensity images for edge detection, and therefore continue to demonstrate high sensitivity to motion blur and low-illumination conditions.

\textbf{Pure event camera-based solutions:}
Event cameras offer strong advantages such as high dynamic range, low latency, and low power consumption. However, the complicated nature of event data demands for novel theories and approaches, and full 6-DoF motion estimation with a single event camera remains a challenging problem. Many works rely on simplifying assumptions. Weikersdorfer et al.~\cite{weikersdorfer2013simultaneous} proposed an event-based 2D SLAM framework for planar motions. Other works rely on a contrast maximization objective that utilizes image-to-image warping, a function that only works if the image transformation is at most a homography (e.g. pure rotation, planar homography)~\cite{gallego2017accurate, stoffregen2019event, gallego2019focus,liu2020globally, peng2021globally}. The first full 6-DoF solution is given by Kim et al.~\cite{kim2016real}, who proposed a complex framework of three decoupled probabilistic filters estimating intensity, depth, and pose, respectively. A geometric solution is given by Rebecq et al.~\cite{rebecq2016evo}, which relies on their earlier ray-density based structure extraction method EMVS~\cite{rebecq2018emvs}. The success of these methods is however limited to small-scale environments and small, dedicated movements on mapping modules. Zhu et al.~\cite{zhu2019neuromorphic} finally present a promising learning-based approach, which however depends on vast amounts of training data, and provides no guarantees of optimality or generality. ESVO uses a stereo event camera~\cite{zhou2021event}, and we compare it against our approach.
    
\textbf{Hybrid event-supported solutions:}
Owing to their difficult nature, event cameras are often combined with other sensors such as IMUs or regular cameras. Censi and Scaramuzza~\cite{censi2014low} present a VO framework that estimates relative poses by fusing events with absolute brightness information. Kueng et al. \cite{kueng2016low} detect features from grayscale images and track the features using the support of event data. Intensity-based methods do not take full advantage of event cameras and may fail due to motion blur in dark or varying illumination settings. While approaches that process images and events individually~\cite{rebecq2017real, vidal2018ultimate} may continue to work if no regular image features are perceived, they still suffer from severe robustness issues if such conditions persist over extended time intervals. Another work that is closely related to ours is introduced by Weikersdorfer et al.~\cite{weikersdorfer2014event}, who extend their previous work~\cite{weikersdorfer2013simultaneous} and include an RGB-D sensor. However, their method is based on an outdated, low-resolution event camera model and relies on a fully voxelized and thus limited-size environment. The accuracy of their probabilistic approach is highly depend on the frequency of depth updates, which limits the speed of the motion.

\section{Depth-Event Visual Odometry}\label{sec:theory}

This section presents the details of our novel stereo depth-event odometry framework, which we denote \textit{DEVO}. We start by seeing an overview of the entire pipeline followed by the details of the event data representation for efficient processing, the semi-dense depth map extraction module, and the final 6 Degree-of-Freedom (DoF) tracking module.

\subsection{Framework overview}

\begin{figure}[t!]
	\centering
	\includegraphics[width=0.85\linewidth]{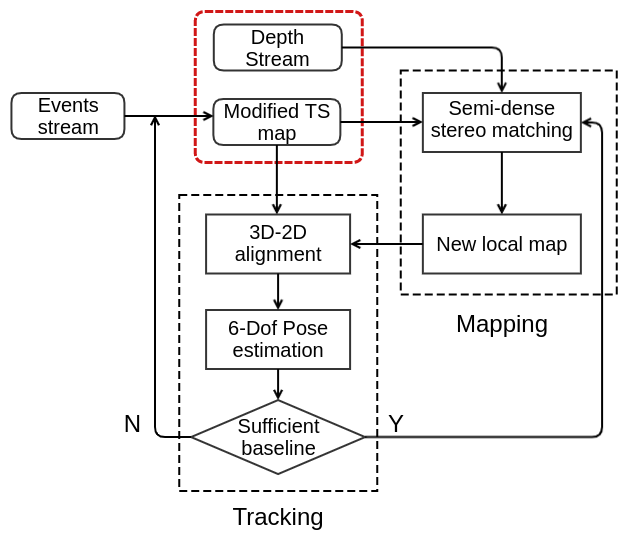}
	\caption{Overview of our proposed \textit{DEVO} visual localization and mapping pipeline for a stereo depth-event system.}
	\label{fig:flowchart}
	\vspace{-0.3cm}
\end{figure}

A flowchart of our proposed method detailing all steps is illustrated in Figure \ref{fig:flowchart}. We start by generating time-surfaces maps which put our event sets into a suitable representation for efficient and accurate edge extraction and alignment. Details are introduced in Section \ref{sec:ts_map}. The representation is used in both a tracking and a mapping module, which--- in analogy to classical visual SLAM architectures---run in two parallel threads. The tracking thread processes only events and incrementally estimates the 6-DoF camera pose by efficient 3D-2D edge alignment. Details of the tracking thread are introduced in Section \ref{sec:6dof_tracking}. The local reference semi-dense depth map is updated at lower framerate inside the mapping thread. It proceeds by extracting the semi-dense edge map from the time-surface maps and assigning depth values from the depth camera readings. The local map is updated whenever sufficient displacements between the current and the reference view has been detected. The operations of the map generation and the reference frame selection strategy are detailed in Section \ref{sec:semidense_mapping}.

\subsection{Event representation}
\label{sec:ts_map}

Let us assume that we are given a set of $N$ events $\mathcal{E} = \{e_{k}\}_{k=1}^{N}$ occurring over a certain time interval. Each event $e_k = \{ \mathbf{x}_k, t_k, b_k \}$ is defined by its image location $\mathbf{x}_k=[\begin{matrix} x_k & y_k \end{matrix}]^T$, timestamp $t_k$, and polarity $b_k$. It is common to not process events asynchronously at the very high rate they occur, but aggregate sets of events accumulated during regularly spaced time intervals into one of three possible representations. The first one is given by space-time volumes of events~\cite{lagorce2016hots}, which are often used in conjunction with accurate but more computationally demanding continuous-time motion representations. The second one is given by simply ignoring the temporal nature of the events, and projecting all events along the temporal dimension onto a virtual binary image in which we then perform feature extraction. Though very efficient, the method re-induces motion blur and requires a careful selection of the time interval length. The third representation is given by time-surface maps (\textbf{TSM}~\cite{lagorce2016hots}), which create an interesting balance between accuracy and efficiency. A \textbf{TSM} is an image in which a high pixel value denotes a recent event. The value at each pixel location $\mathbf{x}$ is a function of an exponential decay kernel and given by
\begin{equation}
    \mathcal{T}(\mathbf{x},t) = \text{exp}(-\frac{t - t_{last}(\mathbf{x})}{\tau}),
\end{equation}
where $t$ is an arbitrary time, and $t_{last}(\mathbf{x}) \leq t$ is the timestamp of the last event triggered at $\mathbf{x}$. $\tau$ denotes the constant decay rate parameter, which requires careful tuning as a function of motion dynamics. The \textbf{TSM} visualizes the history of moving brightness patterns at each pixel location and emphasizes on locations in which motion has been more recent. The values in a \textbf{TSM} are mapped from $[0, 1]$ to $[0, 255]$ for convenient visualization and processing.
We use a modified \textbf{TSM} in which we only consider pixels with a value above a certain threshold $\delta$. Depending on the module (i.e. tracking, or mapping), other pixels are set to 0 or discarded.

\subsection{Mapping module}
\label{sec:semidense_mapping}

The mapping module performs semi-dense point cloud extraction. Let $\mathcal{T}_\text{ref}(\cdot)=\mathcal{T}(\cdot,t_\text{ref})$ be the \textbf{TSM} generated from the set of events $\mathcal{E}$ at time $t_\text{ref}$. The semi-dense region $\mathcal{X}^{\text{ref}}$ for which depth values will be extracted is simply given by all pixels for which the value is larger than $\delta$, i.e. $\mathcal{X}^{\text{ref}} = \{\mathbf{x}\text{ s.t. }\mathcal{T}_\text{ref}(\mathbf{x})>\delta$\}. Based on the assumption that events are pre-dominantly triggered by high-gradient edges in the image, a proper choice of the decay rate $\tau$ and threshold $\delta$ will counteract motion blur and encourage the extracted semi-dense region to align tightly with effective appearance contours.

In order to retrieve the depth value for each point in the semi-dense region, we first warp the depth points from the depth camera at time $t_\text{ref}$ to the event camera. The location in the event camera is given by
\begin{equation}
\mathbf{x}^e_k = \pi_{e}(\mathbf{T}_{ed}\cdot D(z^d_k)\cdot \pi^{-1}_{d}(\mathbf{x}^d_k)),
\end{equation}
where $\pi_{e/d}$ and $\pi^{-1}_{e/d}$ represent the known camera-to-image and the image-to-camera transformations of the event and the depth camera, respectively. They are defined as mapping from the 2D image space to 3D homogeneous space and vice-versa. $D(a)=\text{diag}(a,a,a,1)$ generates a diagonal matrix with elements $a$, $a$, $a$, and $1$ along the diagonal. $\mathbf{T}_{ed}$ is the known $4\times 4$ Euclidean extrinsic transformation matrix from the depth to the event camera. Finally, $\mathbf{x}_k^d$ and $z_k^d$ are a point and its corresponding depth in the depth camera, and $\mathbf{x}_k^e$ is the warped point in the depth frame. The depth $z^e_k$ at the latter point is easily obtained by
\begin{equation}
z^e_k = [0\text{ }0\text{ }1\text{ }0]\cdot\mathbf{T}_{ed}\cdot D(z^d_k)\cdot \pi^{-1}_{d}(\mathbf{x}^d_k),
\end{equation}
Note that the warping maps depth values onto sub-pixel locations rather than event camera pixel centers. It may furthermore induce occlusions or leave pixels with unobserved depths. In order to find a unique depth for each pixel, we create an individual list of nearby warped points from the depth image for each pixel in the semi-dense region. The value of the depth is conditionally set if the pixel is surrounded by warped points from the depth image. A simple depth clustering strategy identifies potential foreground points, and the final value is found by simple interpolation and ray intersection. This ensures that the depth of the pixels in the semi-dense region is always corresponding to foreground points and never affected by occlusions, depth measurement errors, or potential misalignments such as small errors in the extrinsic calibration parameters. The points that have a valid depth assigned to them are renormalized and multiplied by their depth, which finally results in the set $\mathcal{P}^{\text{ref}}$ for our semi-dense 3D point cloud. Note that---in combination with the reference frame poses identified by the tracking module---multiple local maps could be merged into a global map using classical point cloud fusion techniques. The focus of the present work remains however on the localization problem.

\subsection{6-Dof Camera tracking}
\label{sec:6dof_tracking}

\begin{figure}[t!]
	\centering
	\includegraphics[width=\linewidth]{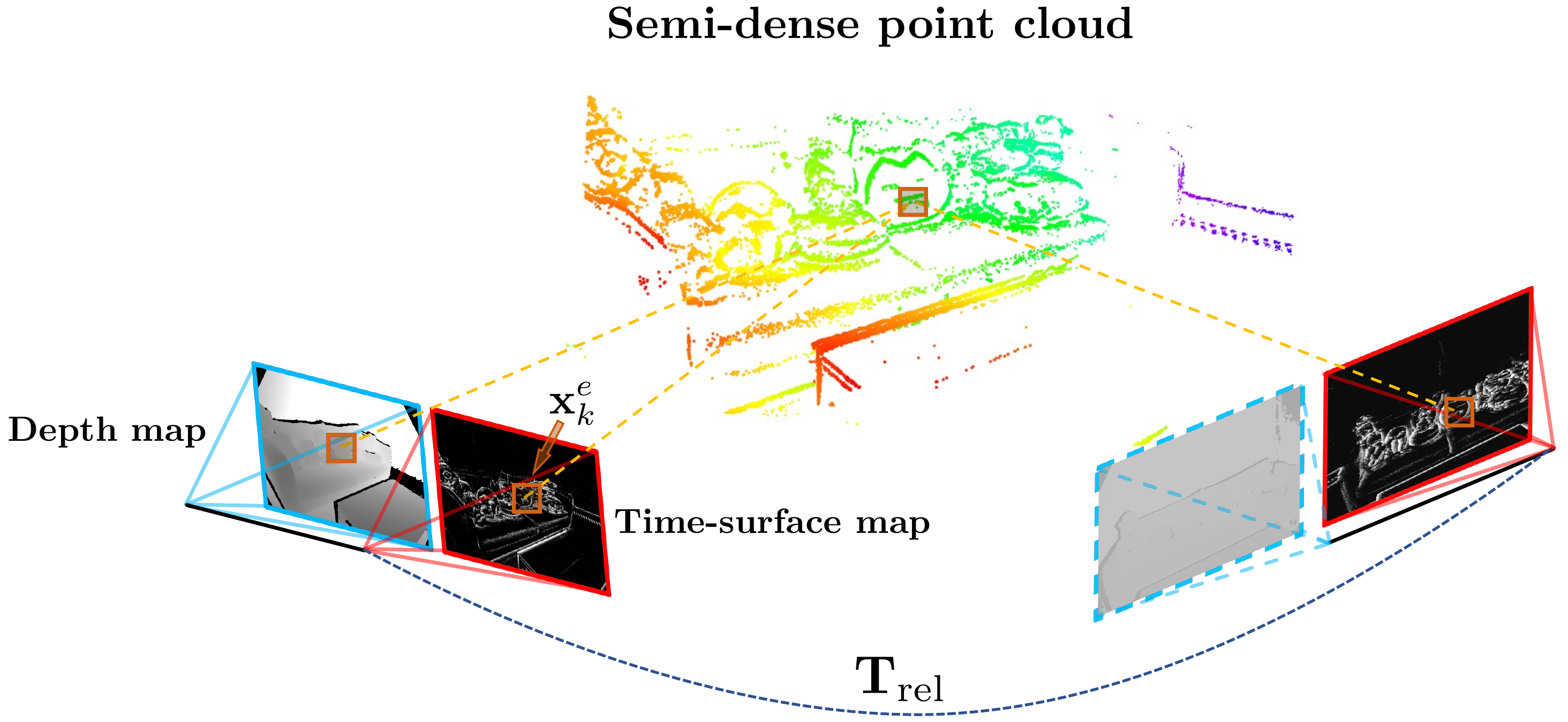}
	\caption{\textit{6-Dof Camera tracking}. Note that the depth image indicated in the dashed frame will not be used in tracking module.}
	\label{fig:6dof_tracking}
	\vspace{-0.3cm}
\end{figure}

With local 3D semi-dense point clouds from the mapping module in hand, we may now proceed to the details of our continuous, 6-DoF motion tracking module. We use the existing event-based localization strategy of Zhou et al. \cite{zhou2021event} in order to align subsequent \textbf{TSM}s with respect to the local semi-dense point cloud. As shown in Figure \ref{fig:6dof_tracking}, the local map (i.e. the reference frame) is furthermore updated by the mapping thread each time the baseline with respect to the previous reference frame exceeds a given threshold. Theoretically, given that events are triggered asynchronously and at very high rate (with temporal resolution in the order of micro-seconds), we could update the pose of the camera with high frequency. Here we choose a rate of 100Hz, which already leads to a strong ability in handling highly dynamic motion.


The tracking proceeds by constructing a potential field in the current view. Based on a hypothesized pose, the reprojected point locations from the semi-dense point cloud then lead to a sampling of this field, and the sum of squares of the sampled values is considered as an energy to be minimized over the pose parameters of the camera. The potential field is constructed by negating and offsetting the \textbf{TSM} at the current time $t_\text{cur}$, i.e. $\overline{\mathcal{T}}_\text{cur}(\cdot) = 1 -\mathcal{T}(\cdot,t_\text{cur})$.

The detailed form of the objective to be minimized is as follows. The local semi-dense 3D point cloud at reference time $t_\text{ref}$ is still given by $\mathcal{P}^{\text{ref}}$. The absolute pose of the current view is given by
\begin{equation}
\mathbf{T}(\boldsymbol{\theta}) =\left[\begin{matrix}\mathbf{R}(\mathbf{q}) & \mathbf{t} \\ \mathbf{0}^\intercal & 1 \end{matrix}\right].
\end{equation}

\noindent where $\boldsymbol{\theta} = [\mathbf{t}^T \text{ } \mathbf{q}^T]^T$ represents a motion parameter vector, $\mathbf{t}$ the position of the camera expressed in a world frame, and $\mathbf{q}$ its orientation as a Rodriguez vector. The relative transformation from the current camera's position to the nearest reference frame is then given as:

\begin{small}
\begin{eqnarray}\small
\mathbf{T}_\text{rel}(\boldsymbol{\theta}_\text{rel}) & =  \mathbf{T}_\text{ref}(\boldsymbol{\theta}_\text{ref})^{-1}\mathbf{T}_\text{cur}(\boldsymbol{\theta}_\text{cur}).\nonumber
\end{eqnarray}
\end{small}
We also define the warping function $W$ that warps a 3D point from the local map to the current frame. It is given by
\begin{equation}
W(\mathbf{x}_k^{\text{ref}};\boldsymbol{\theta}_\text{rel}) = \pi_e(\mathbf{T}^{-1}(\boldsymbol{\theta}_\text{ref})\cdot D(z_k^e)\cdot \pi_e^{-1}(\mathbf{x}_k^{\text{e}})).
\end{equation}
The final goal of the tracking module is to find the optimum motion parameters $\boldsymbol{\theta}_\text{rel}$ that maximize the alignment of the reprojection of the local map $\mathcal{P}^{\text{ref}}$ and the local minima in our current negated \textbf{TSM} $\overline{\mathcal{T}}_\text{cur}(\cdot)$. Using the $W$, the objective function to find the optimum $\boldsymbol{\theta}_\text{rel}$ can be expressed as
\begin{equation}
\mathop{\arg\min}_{\boldsymbol{\theta}_\text{rel}}\sum_{\mathbf{x}_k^{\text{ref}}\in \mathcal{P}^{\text{ref}}}\rho(\overline{\mathcal{T}}_\text{cur}(W(\mathbf{x}_k^{\text{ref}};\boldsymbol{\theta}_\text{rel}))^2),	\label{eq:objectFunc}
\end{equation}
where $\rho$ is a robust loss function. Similar to \cite{zhou2021event}, (\ref{eq:objectFunc}) is reformulated by using a forward compositional Lucas-Kanade method \cite{baker2004lucas}, which refines the incremental motion parameters $\Delta\boldsymbol{\theta}_\text{rel}$ by minimizing:
\begin{equation}
\mathop{\arg\min}_{\boldsymbol{\Delta\theta}_\text{rel}}\sum_{\mathbf{x}_k^{\text{ref}}\in \mathcal{P}^{\text{ref}}}
\rho(\overline{\mathcal{T}}_\text{cur}(W(W(\mathbf{x}_k^{\text{ref}};\Delta\boldsymbol{\theta}_\text{rel});\boldsymbol{\theta}_\text{rel})))^2),
\label{eq:objectFunc2}
\end{equation}
and the new warping function $W(W(\mathbf{x}_k^{\text{ref}};\Delta\boldsymbol{\theta}_\text{rel});\boldsymbol{\theta}_\text{rel})$ is updated in each iteration. The new compositional approach is more efficient than the original method given that the Jacobian of the objective function remains constant at the position of zero increment and can be pre-computed. Smoothness, differentiability and convexity of this method are proven in \cite{zhou2021event}. 

%

\section{Experimental Evaluation}\label{sec:experiments}

We evaluate the performance of our novel visual odometry pipeline on both public and self-collected sequences. We start by introducing further details about the implementation and our hardware configuration. Next, we compare our methods against several alternatives on both mild test cases and more challenging scenarios. Alternatives are given by state-of-the-art event-based and RGB-D or depth-only based approaches. Both qualitative and quantitative results are provided, which demonstrate the effectiveness of our method. We conclude with an analysis of the computational performance of all above mentioned VO systems. 

\begin{figure}[b!]
    \vspace{-0.5cm}
	\centering
	\includegraphics[width=0.8\linewidth]{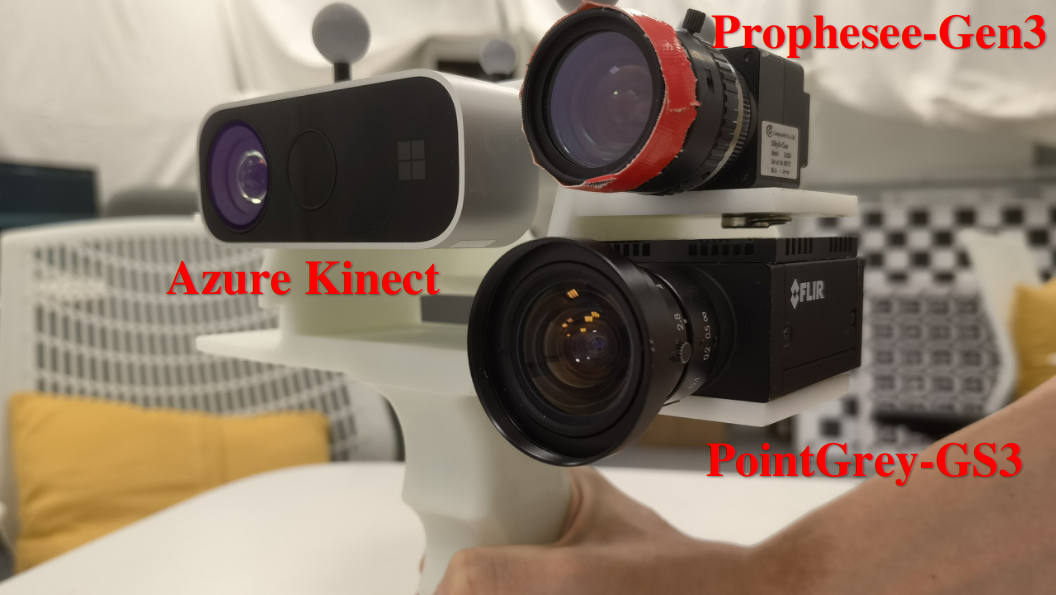}
	\caption{Full sensor system, with event camera, industrial camera and RGB-D sensor.}
	\label{fig:hardware_setup}
\end{figure}

\subsection{Implementation details}
\label{experimental setup and dataset used}

Our first experiments are conducted on the Multi-Vehicle-Stereo-Event-Camera dataset (\textbf{MVSEC}) presented in \cite{2018The}. These publicly available sequences include synchronized event streams, intensity images and depth images with ground truth trajectories. Next, in order to put a full stress test onto all methods, we test the methods on several other, self-collected sequences with different types of textures, motion characteristics, and illumination conditions. For different types of scene textures, the sequences are named \textit{cali}, \textit{table} and \textit{sofa}, respectively. \textit{cali} is a scene with many calibration boards, \textit{table} a standard desktop environment, and \textit{sofa} a living room scene. For each texture, we capture datasets under three different motion speeds, denoted \textit{fast}, \textit{mid} and \textit{slow}. More datasets are captured under a variety of illumination conditions, denoted \textit{bright}, \textit{darkish}, \textit{dim}, \textit{dark}, and \textit{hdr}. All sequences are listed in Table \ref{tab:all comparison}. The sequences are collected by a custom-designed, hardware-synchronized multi-sensor system (cf. Figure \ref{fig:hardware_setup}), which contains a global-shutter industrial camera (PointGrey-GS3), a high-resolution event camera (Prophesee-Gen3), and an RGB-D sensor (Azure Kinect). Detailed specifications are listed in Table \ref{tab:specifications}. The multi-sensor system is intrinsically and extrinsically calibrated, and ground truth for all sequences is captured by a highly accurate external motion capture system.
\begin{table}[t!]
  \vspace{0.2cm}
  \captionsetup{justification=centering}
  \centering
  \caption{Specifications of sensors used in our experiments.}
  \label{tab:Setup}
  \setlength{\tabcolsep}{5.8pt}
    \begin{tabular}{cc|cccccc}
    \toprule
    \multicolumn{2}{c|}{Sensor} & \multicolumn{2}{c}{Exposure Time} & \multicolumn{2}{c}{Resolution} & \multicolumn{2}{c}{Frame Rate} \\
    \midrule
    \multicolumn{2}{c|}{PointGrey-GS3} & \multicolumn{2}{c}{10ms} & \multicolumn{2}{c}{1224$\times$1024} & \multicolumn{2}{c}{30fps} \\
    \multicolumn{2}{c|}{Azure Kinect} & \multicolumn{2}{c}{12.8ms} & \multicolumn{2}{c}{640$\times$576} & \multicolumn{2}{c}{30fps} \\
    \multicolumn{2}{c|}{Prophesee-Gen3} & \multicolumn{2}{c}{-} & \multicolumn{2}{c}{640$\times$480} & \multicolumn{2}{c}{-} \\
    \bottomrule
    \end{tabular}%
  \label{tab:specifications}%
\end{table}
%


\subsection{Comparison against event-based solutions}
\label{compare_with_RGBD}

We first compare our proposed depth-event method \textbf{DEVO} against \textbf{ESVO}, an open-source event-based stereo visual odometry framework published in \cite{zhou2021event}. The two methods are evaluated on the public dataset \textbf{MVSEC} \cite{2018The}. We choose both indoor and outdoor sequences, which are captured by a flying drone inside the room, and a stereo event camera mounted on a vehicle, respectively. Note that the depth measurements in \textbf{MVSEC} are obtained from a LiDAR, which can easily be considered as a replacement for the depth camera in our method.

\begin{figure}[b!]
\centering
\subfigure[Scene]{\includegraphics[width=3.5cm]{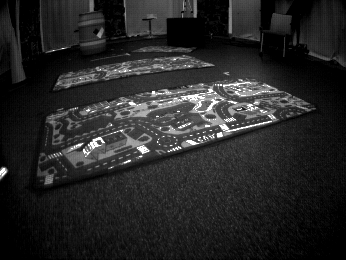}}
\subfigure[Time Surface Map]{\includegraphics[width=3.5cm]{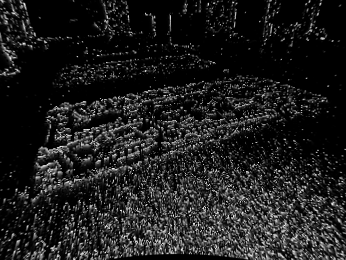}}
\\ 
\centering
\subfigure[Reprojection Map by~\cite{zhou2021event}]{\includegraphics[width=3.5cm]{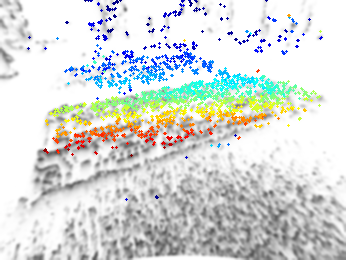}}
\subfigure[Reprojection Map (Ours)]{\includegraphics[width=3.5cm]{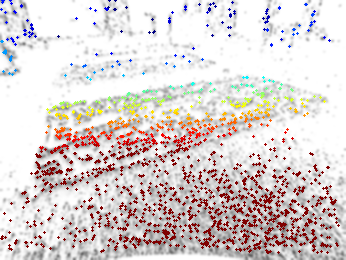}}
\caption{\label{MVSEC_figure_comparison}Visualization of an image (top left) and \textbf{TSM} (top right) from sequence \textit{indoor flying2} and corresponding reprojections into a nearby frame for \textbf{ESVO} (bottom left) and \textbf{DEVO} (bottom right). The coloring indicates the depth of each point.}
\end{figure}

Quantitative results are listed in Table \ref{tab:ESVO Comparison}. As can be observed, \textbf{DEVO} clearly outperforms \textbf{ESVO} in all sequences. It should be noted that sequences \textit{indoor flying2} and \textit{indoor flying4} are much more challenging than the other two sequences owing to high noise in the event streams caused by a combination of difficult texture and highly dynamic platform motion. Examples of the sequence are indicated in Figure \ref{MVSEC_figure_comparison}. The ground surface triggers a large number of noisy events for which depth is hard to observe. This severely influences the mapping result of \textbf{ESVO} and--- due to the highly interleaved tracking and mapping modules---causes tracking failures in this fast exploration scenario. In contrast, \textbf{DEVO} directly reads the depth from the depth sensor, the quality of which is less influenced by the noisy nature of the texture. The independent depth readings significantly contribute to the robustness of the entire system when the inputs of the event camera degrade. Furthermore, the ability of any stereo method to perceive depths beyond a certain range is limited by the baseline of the system, which is why \textbf{ESVO} is unable to provide competitive results on the outdoor sequence.

\subsection{Comparison against RGB-D camera-based solutions}
\label{robustness comparision}

\begin{table}[t!]
\vspace{0.2cm}
\captionsetup{justification=centering}
\centering
  \caption{Comparison on \textbf{MVSEC} Datasets\\ $\left[\mathbf{R}_\text{\upshape rpe}\text{\upshape : °/s}, \mathbf{t}_\text{\upshape rpe}\text{\upshape : cm/s}, \mathbf{t}_\text{\upshape ate}\text{\upshape : cm}\right]$}
  \label{tab:ESVO Comparison}
  \renewcommand\arraystretch{1.3}
 \setlength{\tabcolsep}{3.7pt}
\begin{tabular}{llccccccc}
\toprule
                              &  & \multicolumn{3}{c}{\textbf{DEVO}}              &  & \multicolumn{3}{c}{\textbf{ESVO}} \\ \cline{3-5} \cline{7-9} 
\textit{Sequence}             &  & $\mathbf{R}_\text{rpe}$          & $\mathbf{t}_\text{rpe}$          & $\mathbf{t}_\text{ate}$           &  & $\mathbf{R}_\text{rpe}$       & $\mathbf{t}_\text{rpe}$       & $\mathbf{t}_\text{ate}$      \\ \cline{1-5} \cline{7-9} 
\textit{upenn indoor flying1} &  & \textbf{0.30} & \textbf{0.88} & \textbf{20.58} &  & 0.37      & 1.63      & 21.68     \\
\textit{upenn indoor flying2} &  & \textbf{0.36} & \textbf{1.12} & \textbf{11.33} &  & -         & -         & -         \\
\textit{upenn indoor flying3} &  & \textbf{0.53} & \textbf{1.21} & \textbf{10.60} &  & 0.54      & 2.14      & 25.40     \\
\textit{upenn indoor flying4} &  & \textbf{0.53} & \textbf{1.44} & \textbf{13.16} &  & -         & -         & -         \\
\textit{upenn outdoor day1}   &  & \textbf{0.30} & \textbf{7.77} & \textbf{88.70} &  & -         & -         & -         \\ \bottomrule
\end{tabular}
\end{table}

\begin{table}[b!]
  \vspace{-0.2cm}
  \centering
  \caption{Comparison for different light conditions}
  \label{tab:MPL Availability}
  \setlength{\tabcolsep}{13pt}
    \begin{tabular}{cc|cccccccc}
    \toprule
    \multicolumn{2}{c|}{Sequence} & \multicolumn{2}{c}{\textbf{DEVO}} & \multicolumn{2}{c}{\textbf{KinectFusion}} & \multicolumn{2}{c}{\textbf{Canny-VO}}  \\
    \midrule
    \multicolumn{2}{c|}{\textit{light}} & \multicolumn{2}{c}{\textcolor[rgb]{ 0,  .69,  .314}{\Checkmark}} & \multicolumn{2}{c}{\textcolor[rgb]{ 0,  .69,  .314}{\Checkmark}} & \multicolumn{2}{c}{\textcolor[rgb]{ 0,  .69,  .314}{\Checkmark}} \\
    \multicolumn{2}{c|}{\textit{darkish}} & \multicolumn{2}{c}{\textcolor[rgb]{ 0,  .69,  .314}{\Checkmark}} & \multicolumn{2}{c}{\textcolor[rgb]{ 0,  .69,  .314}{\Checkmark}} & \multicolumn{2}{c}{\textcolor[rgb]{ 1,  0,  0}{\XSolidBrush}}  \\
    \multicolumn{2}{c|}{\textit{dim}} & \multicolumn{2}{c}{\textcolor[rgb]{ 0,  .69,  .314}{\Checkmark}} & \multicolumn{2}{c}{\textcolor[rgb]{ 0,  .69,  .314}{\Checkmark}} & \multicolumn{2}{c}{\textcolor[rgb]{ 1,  0,  0}{\XSolidBrush}}  \\
    \multicolumn{2}{c|}{\textit{dark}} & \multicolumn{2}{c}{\textcolor[rgb]{ 0,  .69,  .314}{\Checkmark}} & \multicolumn{2}{c}{\textcolor[rgb]{ 0,  .69,  .314}{\Checkmark}} & \multicolumn{2}{c}{\textcolor[rgb]{ 1,  0,  0}{\XSolidBrush}} \\
    \multicolumn{2}{c|}{\textit{HDR}} & \multicolumn{2}{c}{\textcolor[rgb]{ 0,  .69,  .314}{\Checkmark}} & \multicolumn{2}{c}{\textcolor[rgb]{ 0,  .69,  .314}{\Checkmark}} & \multicolumn{2}{c}{\textcolor[rgb]{ 1,  0,  0}{\XSolidBrush}} \\
    \bottomrule
    \end{tabular}%
  \label{tab:addlabel}%
\end{table}%

\begin{table*}[t!]
\vspace{0.2cm}
\caption{Relative pose error and absolute trajectory error on self-collected datasets $\left[\mathbf{R}_\text{\upshape rpe}\text{\upshape : °/s}, \mathbf{t}_\text{\upshape rpe}\text{\upshape : cm/s}, \mathbf{t}_\text{\upshape ate}\text{\upshape : cm}\right]$}
  \centering
  \label{tab:all comparison}
  \setlength{\tabcolsep}{9.7pt}
  \renewcommand\arraystretch{1.2}
\begin{tabular}{lcccccccccccc}
\toprule
                             & \multicolumn{1}{c}{} & \multicolumn{3}{c}{\textbf{DEVO}}                      & \multicolumn{1}{c}{} & \multicolumn{3}{c}{\textbf{Canny-VO}}                   &  & \multicolumn{3}{c}{\textbf{KinectFusion}}              \\ \cline{3-5} \cline{7-9} \cline{11-13} 
\textit{Sequence}            &                       & $\mathbf{R}_\text{rpe}$          & $\mathbf{t}_\text{rpe}$          & $\mathbf{t}_\text{ate}$           &                       & $\mathbf{R}_\text{rpe}$          & $\mathbf{t}_\text{rpe}$          & $\mathbf{t}_\text{ate}$           &  & $\mathbf{R}_\text{rpe}$          & $\mathbf{t}_\text{rpe}$          & $\mathbf{t}_\text{ate}$           \\ \cline{1-5} \cline{7-9} \cline{11-13} 
\textit{cali\_bright\_fast}  &                       & \textbf{3.73} & 2.03          & 23.67          &                       & 3.81          & \textbf{1.47} & 15.55          &  & 3.74          & 1.81          & \textbf{15.34} \\
\textit{cali\_bright\_mid}   &                       & 1.44          & 1.77          & \textbf{16.90} &                       & 1.42          & \textbf{1.26} & 21.23          &  & \textbf{1.35} & 1.77          & 19.22          \\
\textit{cali\_bright\_slow}  &                       & \textbf{0.97} & 0.78          & 11.85          &                       & 1.03          & \textbf{0.59} & \textbf{7.16}  &  & 0.99          & 0.89          & 14.52          \\
\textit{cali\_darkish\_slow} &                       & 1.03          & \textbf{0.91} & 18.02          &                       & -             & -             & -              &  & \textbf{1.02} & 0.93          & \textbf{11.34} \\
\textit{cali\_dim\_slow}     &                       & \textbf{1.55} & 0.88          & 35.38          &                       & -             & -             & -              &  & 1.62          & \textbf{0.82} & \textbf{9.05}  \\
\textit{cali\_dark\_fast}    &                       & \textbf{0.58} & \textbf{0.87} & 26.43          &                       & -             & -             & -              &  & 0.63          & 0.92          & \textbf{12.61} \\
\textit{cali\_dark\_mid}     &                       & \textbf{0.49} & 0.60          & 17.65          &                       & -             & -             & -              &  & 0.54          & \textbf{0.59} & \textbf{12.89} \\
\textit{cali\_dark\_slow}    &                       & \textbf{0.24} & 0.31          & 9.85           &                       & -             & -             & -              &  & 0.26          & \textbf{0.23} & \textbf{8.97}  \\
\textit{cali\_hdr\_slow}     &                       & \textbf{0.92} & 0.79          & 21.55          &                       & -             & -             & -              &  & 0.95          & \textbf{0.71} & \textbf{11.10} \\
\textit{table\_bright\_fast} &                       & 1.50          & 2.42          & 46.37          &                       & 1.51          & \textbf{2.22} & 30.81          &  & \textbf{1.38} & 2.75          & \textbf{27.00} \\
\textit{table\_bright\_mid}  &                       & 1.16          & 1.53          & 27.83          &                       & 1.18          & \textbf{1.25} & \textbf{19.68} &  & \textbf{1.10} & 1.79          & 21.71          \\
\textit{table\_bright\_slow} &                       & 0.63          & 1.00          & \textbf{19.5}  &                       & 0.64          & \textbf{0.82} & 26.94          &  & \textbf{0.59} & 1.15          & 22.48          \\
\textit{sofa\_bright\_fast}  &                       & \textbf{2.60} & 2.30          & 30.22          &                       & 2.63          & \textbf{1.90} & \textbf{23.89} &  & 2.61          & 3.64          & 27.79          \\
\textit{sofa\_bright\_mid}   &                       & 5.28          & 4.02          & \textbf{13.4}  &                       & \textbf{1.18} & \textbf{1.25} & 19.68          &  & 3.13          & 7.62          & 71.5           \\
\textit{sofa\_bright\_slow}  &                       & 1.47          & 1.16          & \textbf{10.94} &                       & \textbf{0.64} & \textbf{0.82} & 26.94          &  & 1.47          & 1.21          & 21.82          \\ \bottomrule
\end{tabular}
\end{table*}

In order to analyze robustness under challenging illumination conditions, we compare our method against two classical approaches that rely on RGB-D cameras or depth sensors, only. They are given by \textbf{KinectFusion}~\cite{newcombe2011kinectfusion} and \textbf{Canny-VO}~\cite{zhou2018canny}. We apply all methods to our self-collected datasets. We conduct three types of experiments, and all absolute trajectory errors (ATE) and relative pose errors (RPE) are summarized in Table \ref{tab:all comparison}:
\begin{itemize}
    \item \textit{Variation of light conditions}: We apply all methods on a series of sequences with different illumination conditions denoted \textit{bright}, \textit{darkish}, \textit{dim}, \textit{dark} and \textit{high dynamic range (hdr)}. As summarized in Table~\ref{tab:MPL Availability}, both \textbf{DEVO} and \textbf{KinectFusion} are able to continuously track through all sequences, while \textbf{Canny-VO} proves to be fragile when applied in poor illumination conditions. The reason is a lack of edge features caused by blur and poor contrast in dark scenarios. 
    \item \textit{Variation of motion characteristics}: We evaluate the performance of all methods for different motion dynamics. The sequences are denoted \textit{fast}, \textit{mid}, or \textit{slow} to indicate the different camera dynamics. As can be observed in Table~\ref{tab:all comparison}, all methods have a remarkable ability to handle dynamic scenarios for standard depth camera frame rate.
    \item \textit{Variation of depth camera frame rate}: In order to analyse each method's ability to operate in an energy-saving mode, we finally test all methods for different depth camera frame rates between 30Hz and 1Hz in the \textit{table} environment and for three different camera dynamics. As indicated in Table \ref{tab:Frequency}, only our method is able to maintain stable tracking for all depth camera frame rates down to 1Hz. While accuracy decreases for more agile motion, it should be noted that the motion on these sequences is highly aggressive. 
\end{itemize}

\begin{table}[ht!]
\captionsetup{justification=centering}
  \centering
  \caption{Comparison for different depth frame rates\\ $\left[\mathbf{R}_\text{\upshape rpe}\text{\upshape : °/s}, \mathbf{t}_\text{\upshape rpe}\text{\upshape : cm/s}, \mathbf{t}_\text{\upshape ate}\text{\upshape : cm}\right]$}
  \label{tab:Frequency}
  \setlength{\tabcolsep}{1.9pt}
    \renewcommand\arraystretch{1.25}
\begin{tabular}{ccccccccccccc}
\toprule
\textit{Frequency} & \multicolumn{1}{l}{} & \multicolumn{3}{c}{\textbf{DEVO}}                & \multicolumn{1}{l}{\textbf{}} & \multicolumn{3}{c}{\textbf{Canny-VO}} & \multicolumn{1}{l}{\textbf{}} & \multicolumn{3}{c}{\textbf{KinectFusion}} \\ \cline{3-5} \cline{7-9} \cline{11-13} 
\textit{Fast}      &                      & $\mathbf{R}_\text{rpe}$           & $\mathbf{t}_\text{rpe}$            &  $\mathbf{t}_\text{ate}$          &                               & $\mathbf{R}_\text{rpe}$ & $\mathbf{t}_\text{rpe}$          & $\mathbf{t}_\text{ate}$            &                               & $\mathbf{R}_\text{rpe}$           & $\mathbf{t}_\text{rpe}$    & $\mathbf{t}_\text{ate}$            \\ \cline{1-5} \cline{7-9} \cline{11-13} 
30                 &                      & 1.50           & 2.42           & 46.37          &                               & 1.51 & \textbf{2.22} & 30.81          &                               & \textbf{1.37}   & 2.75   & \textbf{27.00}  \\
15                 &                      & \textbf{2.92}  & 4.86           & 46.70          &                               & 2.96 & \textbf{4.46} & \textbf{29.11} &                               & 3.04            & 8.36   & 37.99           \\
10                 &                      & \textbf{4.26}  & 7.32           & 47.92          &                               & 4.65 & \textbf{6.55} & \textbf{34.68} &                               & 4.83            & 17.33  & 66.78           \\
5                  &                      & \textbf{7.73}  & \textbf{14.73} & 57.89          &                               & -    & -             & -              &                               & 9.01            & 26.86  & \textbf{55.09}  \\
1                  &                      & \textbf{18.58} & \textbf{49.16} & \textbf{76.04} &                               & -    & -             & -              &                               & -               & -      & -               \\ \cline{1-5} \cline{7-9} \cline{11-13} 
\textit{Medium}    &                      & $\mathbf{R}_\text{rpe}$          & $\mathbf{t}_\text{rpe}$            & $\mathbf{t}_\text{ate}$          &                               & $\mathbf{R}_\text{rpe}$  & $\mathbf{t}_\text{rpe}$           & $\mathbf{t}_\text{ate}$            &                               & $\mathbf{R}_\text{rpe}$             & $\mathbf{t}_\text{rpe}$   & $\mathbf{t}_\text{ate}$             \\ \cline{1-5} \cline{7-9} \cline{11-13} 
30                 &                      & 1.16           & 1.53           & 27.83          &                               & 1.18 & \textbf{1.25} & \textbf{19.68} &                               & \textbf{1.10}   & 1.79   & 21.71           \\
15                 &                      & 2.29           & 3.01           & 24.20          &                               & 2.34 & \textbf{2.48} & \textbf{20.08} &                               & \textbf{2.17}   & 3.56   & 21.31           \\
10                 &                      & 3.39           & 4.51           & 21.46          &                               & 3.49 & \textbf{3.70} & \textbf{20.34} &                               & \textbf{3.20}   & 5.58   & 58.73           \\
5                  &                      & \textbf{6.55}  & \textbf{9.20}  & \textbf{21.55} &                               & -    & -             & -              &                               & 7.16            & 15.80  & 37.07           \\
1                  &                      & \textbf{18.46} & \textbf{35.24} & \textbf{51.93} &                               & -    & -             & -              &                               & -               & -      & -               \\ \cline{1-5} \cline{7-9} \cline{11-13} 
\textit{Slow}      &                      & $\mathbf{R}_\text{rpe}$         &$\mathbf{t}_\text{rpe}$          & $\mathbf{t}_\text{ate}$           &                               & $\mathbf{R}_\text{rpe}$ & $\mathbf{t}_\text{rpe}$          & $\mathbf{t}_\text{ate}$           &                               & $\mathbf{R}_\text{rpe}$            & $\mathbf{t}_\text{rpe}$   & $\mathbf{t}_\text{ate}$            \\ \cline{1-5} \cline{7-9} \cline{11-13} 
30                 &                      & 0.63           & 1.00           & \textbf{19.50} &                               & 0.64 & \textbf{0.82} & 26.94          &                               & \textbf{0.59}   & 1.15   & 22.48           \\
15                 &                      & 1.21           & 1.97           & \textbf{18.44} &                               & 1.24 & \textbf{1.65} & 26.01          &                               & \textbf{1.13}   & 2.29   & 22.22           \\
10                 &                      & 1.76           & 2.98           & \textbf{18.46} &                               & 1.82 & \textbf{2.46} & 26.32          &                               & \textbf{1.66}   & 3.41   & 22.10           \\
5                  &                      & \textbf{3.28}           & 6.09           & \textbf{17.61} &                               & 3.71 & \textbf{5.54} & 27.97          &                               & -   & -  & -          \\
1                  &                      & \textbf{10.66} & \textbf{31.15} & \textbf{37.88} &                               & -    & -             & -              &                               & -               & -      & -               \\ \bottomrule
\end{tabular}
\vspace{-0.2cm}
\end{table}

\subsection{Computational Performance}
\label{computational performance}

As can be observed from the ATE and RPE errors listed in Table~\ref{tab:all comparison}, \textbf{DEVO} has comparable performance with other state-of-the-art methods from the literature. Although \textbf{Canny-VO} demonstrates lowest RPE errors in good lighting conditions, it shows degrading performance when the illumination becomes more challenging. We perceive \textbf{KinectFusion} as the strongest competitor of our method as it achieves comparable accuracy on all sequences. However, it should be noted that \textbf{KinectFusion} results have been obtained by putting the software into a high-performance setting that requires sufficient computing power to run. \textbf{KinectFusion} results in this paper have been obtained by a 32 core CPU and two Nvidia RTX 2080Ti. By comparison, the other two methods run on an 8 core CPU, only. It should furthermore be noted that \textbf{KinectFusion} depends on sufficiently high depth camera framerates, which again induces larger energy consumption.


\section{Conclusion}

We present a novel approach to visual odometry that relies on a stereo depth-event camera. In comparison to depth-only alternatives, it handles faster motion and works more efficiently by requiring lower depth image frame rates and by performing semi-dense image processing. In comparison to RGB-D visual odometry solutions, it successfully handles challenging or low illumination scenarios. In summary, our proposed method handles a large spectrum of challenging situations, and we believe that it could represent a highly interesting approach for intelligent mobile systems that require indoor localization. A pre-condition would however be that event cameras are becoming more affordable in the future. 


{\small
\bibliographystyle{IEEEtran}
\bibliography{root}
}

\end{document}